\documentclass[letterpaper, 10 pt, conference]{ieeeconf}
\usepackage{graphicx}
\usepackage{float}
\usepackage{hyperref}
\usepackage{caption}
\usepackage{subcaption}
\usepackage{hyperref}
\usepackage{changes}
\usepackage{euscript}[mathcal]
\usepackage[ruled,linesnumbered]{algorithm2e}
\usepackage{tikz}

\usepackage{amsmath}
\usepackage{amsfonts}
\usepackage{amsthm}
\SetKwBlock{Repeatforever}{repeat forever}{}
\SetKwProg{Repeat2}{repeat}{}{end}

\IEEEoverridecommandlockouts 
\theoremstyle{definition}

\newcommand{\refappendix}[1]{\hyperref[#1]{Appendix~\ref*{#1}}}
\begin{document}
\SetEndCharOfAlgoLine{}

\title{MAMBPO: Sample-efficient multi-robot reinforcement learning using learned world models}
\author{Dani{\"e}l Willemsen$^{1}$, Mario Coppola$^{1}$ and Guido C.H.E. de Croon$^{1}$
\thanks{$^{1}$ MAVLab, Control \& Operations Department, Faculty of Aerospace Engineering, Delft University of Technology, 2628 HS Delft, The Netherlands. 
{\tt\small j.daniel.willemsen@gmail.com}\newline%
{\tt\small m.coppola@tudelft.nl}\newline
{\tt \small  g.c.h.e.decroon@tudelft.nl}}%
}
\maketitle
\thispagestyle{plain}
\pagestyle{plain}

\begin{abstract}
Multi-robot systems can benefit from reinforcement learning (RL) algorithms that learn behaviours in a small number of trials, a property known as sample efficiency. This research thus investigates the use of learned world models to improve sample efficiency. We present a novel multi-agent model-based RL algorithm: \textit{Multi-Agent Model-Based Policy Optimization (MAMBPO)}, utilizing the Centralized Learning for Decentralized Execution (CLDE) framework. CLDE algorithms allow a group of agents to act in a fully decentralized manner after training. This is a desirable property for many systems comprising of multiple robots. 
MAMBPO uses a learned world model to improve sample efficiency compared to model-free Multi-Agent Soft Actor-Critic (MASAC). We demonstrate this on two simulated multi-robot tasks, where MAMBPO achieves a similar performance to MASAC, but requires far fewer samples to do so. Through this, we take an important step towards making real-life learning for multi-robot systems possible.
\end{abstract}

\section{Introduction}
Reinforcement Learning (RL) for multi-robot systems in the real world can be prohibitively expensive due to the large amount of time required, as well as the possibility of breaking hardware from trial and error of the algorithm \cite{kober_reinforcement_2013}.
Training in a simulation is possible, but it introduces a reality-gap, meaning that the performance may not transfer to the real world due to simulator imperfections. 
Therefore, to (re-)train in the real-world, there is a need for sample-efficient learning algorithms, i.e., algorithms that can learn behaviors with a small number of trials.

\begin{figure}[t]
    \centering
    \includegraphics[width=0.4\textwidth]{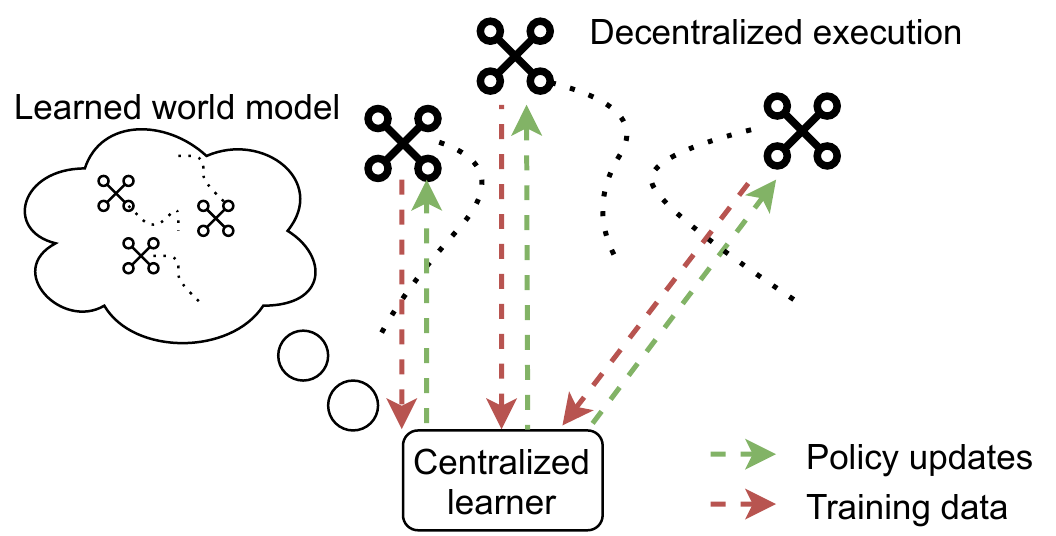}
    \caption{Visualization of MAMBPO, with the CLDE framework, in a real life scenario. Drones act fully decentralized, but share their experiences with a centralized learner when possible. The centralized learner learns a world model and uses this world model to generate additional training data, which is then used to update the robots' policies.}
    \label{fig:overview}
\end{figure}

This research proposes the use of model-based learning to improve sample efficiency of multi-agent RL. 
The idea is to learn a model of the world dynamics to generate additional training data, speeding up the learning process.
To do this, we introduce a novel model-based multi-agent RL algorithm: \emph{Multi-Agent Model-Based Policy Optimization (MAMBPO)}, which is a multi-agent adaptation of the Model-Based Policy Optimization (MBPO) algorithm \cite{janner_when_2019}. We demonstrate MAMBPO on two multi-agent domains, showing that it is more sample-efficient than state-of-the-art model-free approaches.
MAMBPO adopts the \emph{Centralized Learning for Decentralized Execution (CLDE)} framework.
The CLDE framework allows centralized information to be used during training to improve the stability and performance of the algorithms. Then, during execution, no centralized information is required and all agents act fully decentralized, a desirable characteristic in many multi-robot systems \cite{sahin_swarm_2008}.
A diagram of the framework is shown in \autoref{fig:overview}.
To the best of our knowledge, this is the first application of learning with a generative world model within the CLDE framework.  

This paper is structured as follows. \autoref{sec:rel} details related works and establishes the novelty of our approach. \autoref{sec:method} introduces MAMBPO, explaining how it combines CLDE learning and model-based RL. In \autoref{sec:experiments}, MAMBPO is tested on two benchmark domains: cooperative navigation and cooperative predator-prey. 
\autoref{sec:conclusion} discusses the main conclusions and  future work towards real-life learning for multi-robot systems.

\section{Related Work}
\label{sec:rel}

Various CLDE algorithms for continuous action spaces have been developed. 
Multi-Agent Deep Deterministic Policy Gradient (MADDPG) is a CLDE variant of the single-agent DDPG algorithm \cite{lowe_multi-agent_2017}.
MADDPG, suitable for cooperative, competitive, and mixed domains, uses centralized critics that take the concatenated joint observations and actions of all agents as an input and outputs the estimated value for a single agent.
The actors are decentralized, taking only the agents' own observations as an input and outputting a deterministic action. 
Iqbal and Sha proposed Multi-Actor-Attention-Critic (MAAC) \cite{iqbal_actor-attention-critic_2019}, using an attention mechanism to improve the scalability.
Another alternative is a maximum entropy variant called Multi-Agent Soft Actor-Critic (MASAC). 
In maximum entropy RL, agents have an incentive to maximize exploration within the policy. This can increase both stability as well as sample efficiency \cite{iqbal_actor-attention-critic_2019, gupta_probabilistic_2019}. MASAC will serve as our model-free benchmark, and is integrated for policy optimization in our model-based algorithm. 

Model-based learning has recently shown great sample-efficiency improvements for  single-agent domains. Early model-based RL approaches suffer from poor asymptotic performance due to a phenomenon known as model bias, where agents exploit inaccuracies in a model, resulting in suboptimal policies in the real environment. Recent work utilizes models that explicitly and simultaneously take into account aleatoric and epistemic uncertainty, for example through the use of ensembles of stochastic neural network models \cite{chua_deep_2018}, reducing the impact of model bias.
Such models have resulted in algorithms that are comparable to state-of-the-art model-free algorithms in terms of asymptotic performance, yet are able to learn with up to an order of magnitude greater sample efficiency \cite{janner_when_2019}. 

Few works combine CLDE and model-based learning.
Krupnik et al. \cite{krupnik_multi-agent_2019} proposed the use of Model-Predictive Control based on the learned world model.
However, this requires using the model during execution, which in turn requires centralized information, making it unsuitable for decentralized execution.
Other research investigates the possibility of using a model to not only predict the environment but also predict the policies of opposing agents in 2-agent competitive domains \cite{indarjo_deep_nodate}. This approach, however, was only evaluated on a task where the opponent agent has a fixed policy, rather than being an opponent that is learning concurrent to the agent itself, which then does not address non-stationarity.

\section{Methodology}
\label{sec:method}
In this section, we present our RL method termed MAMBPO.
\autoref{subsec:notation} describes the problem and relevant notation.
\autoref{subsec:mambpo} explains MAMBPO and its two main components: policy optimization through Multi-Agent Soft Actor-Critic (MASAC) and model-based learning through Model-Based Policy Optimization (MBPO).

\subsection{Problem Description \& Notation}
\label{subsec:notation}
\begin{figure}[h!]
    \centering
    \includegraphics[width=0.45\textwidth]{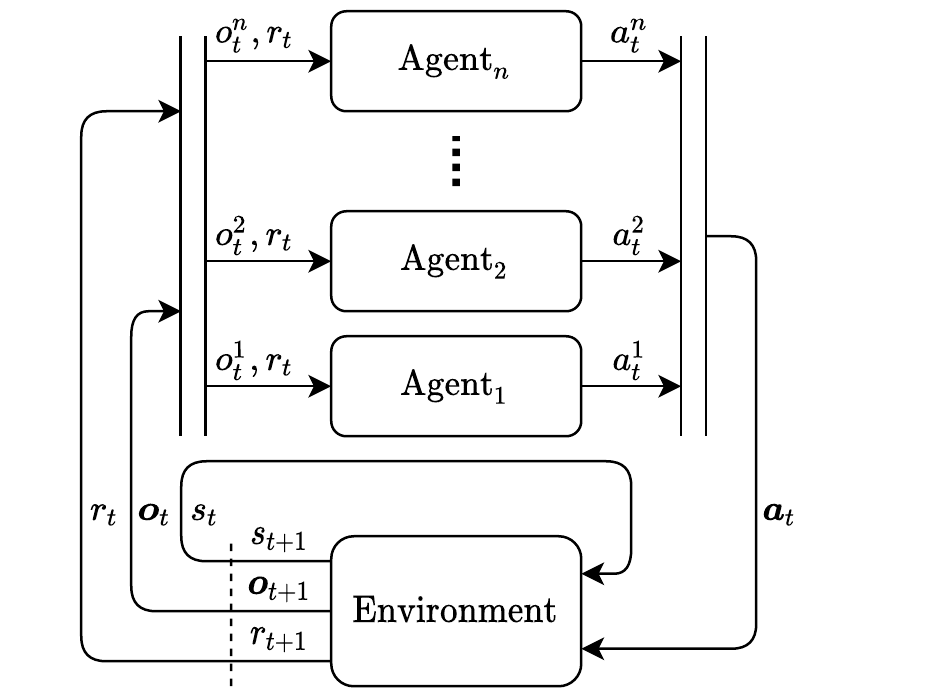}
    \caption{Description of the interaction between the agents and the environment in a Dec-POMDP.}
    \label{fig:dec-pomdp}
\end{figure}

In this research, we consider the problem of a group of agents interacting cooperatively within an environment, formalized in the Decentralized Partially Observable Markov Decision Process (Dec-POMDP) framework \cite{oliehoek_decentralized_2012}. This framework is illustrated in \autoref{fig:dec-pomdp}.  At every timestep $t$, every agent $i \in \left\{1, ..., n\right\}$ simultaneously receives an individual observation $o_t^i$ from a joint observation $\boldsymbol{o}_t := \left\{o_t^i \right\}_{i=1}^n$. An action $a_t^i$ is then sampled from a (stochastic) policy ($\pi^i$), such that $a_t^i \sim \pi^i(\cdot |o_t^i)$. We use the $\pi^i(\cdot|o_t^i)$ notation to indicate the full probability distribution of $\pi^i$ and we use the notation $\pi^i(a_t^i|o_t^i)$ to indicate the probability density function at a particular value $a_t^i$. The concatenation of the actions from all agents is the joint action, denoted $\boldsymbol{a}_t := \left\{a_t^i \right\}_{i=1}^n$.  Note that we use superscripts $^i$ to indicate individual observations ($o_t^i$) and actions ($a_t^i$) rather than the joint observation ($\boldsymbol{o}_t$) and action ($\boldsymbol{a}_t$), which are indicated in bold and without a superscript. Similarly, we use $\boldsymbol{\pi}$ to indicate the joint policy of all agents. The environment takes the joint action and a hidden environment state $s_{t}$ to (stochastically) produce a new hidden state through sampling from the state-transition function $P$, such that $s_{t+1} \sim P(\cdot | s_t, \boldsymbol{a} _t)$. The environment then produces a joint observation and reward through the observation function $O$ and reward function $R$ respectively\footnote{We define the reward $r_{t+1}$ as the reward received after taking action $\boldsymbol{a}_{t}$ and is received by the agents simultaneously with $\boldsymbol{o}_{t+1}$. This notation differs from some literature that would denote this reward with $r_t$ as it is the reward that is a direct consequence of the action taken on time step $t$.}: $\boldsymbol{o}_{t+1} \sim O(s_{t+1})$, $r_{t+1} \sim R(s_{t+1})$. The goal for each agent is to find policy $\pi^i$ that maximizes the expected return, which is the expected value of the cumulative reward, discounted with some factor $\gamma$ over time until some terminal condition is reached:  $\mathbb{E}[G_k] = \mathbb{E}[\sum_{t={k}}^{t=t_{terminal}} \gamma^{k-t} r_{t+1}]$. The state-action value $Q$ is a measure for the expected return of an agent given a certain starting state $s$, joint policy $\pi$ and initial joint action $\boldsymbol{a}$: $Q^{\pi}(s,\boldsymbol{a}) = \mathbb{E}_{\pi}\left[\sum_{t={k}}^{t=t_{terminal}} \gamma^{k-t} r_{t+1} \middle| s_k = s, \boldsymbol{a}_k = \boldsymbol{a} \right]$.

A Dec-POMDP can either have finite sets of states, actions, and observations, or it can have continuous state-, action-, and observation-spaces. In this article, we consider Dec-POMDPs of the continuous type.
\theoremstyle{definition}

\subsection{Multi-Agent Model-Based Policy Optimization}
\label{subsec:mambpo}
Our novel MAMBPO algorithm is a combination of model-based RL and CLDE. The core idea is that it is easier for an agent to learn a model of the world than it is to optimize a policy. Therefore, MAMBPO learns a world model and then uses this world model to generate additional experience to train the agent's policy on. We train the policy through  MASAC, a CLDE variant of Soft Actor-Critic \cite{iqbal_actor-attention-critic_2019, gupta_probabilistic_2019}. This two-step process of model learning and policy optimization allows for improved sample efficiency compared to directly optimizing a policy. A high level overview of the algorithm can be seen in \autoref{alg:mambpo}. Agents interact with the environment and a centralized replay buffer ($\EuScript{B}_{env}$) is used to store the transitions. Periodically, a centralized world model ($\hat{p}_{\theta}$) is trained on this replay buffer. The world model is then used to generate additional samples, stored in a separate replay buffer ($\EuScript{B}_{model}$). We supplement this second replay buffer with a small amount (in our case 10\%) of real environment data, similar to the original MBPO implementation \cite{janner_when_2019}. The second replay buffer, $\EuScript{B}_{model}$, is used to train the actor and critic networks. The key insight that allows this algorithm to perform sample-efficient learning is that the larger diversity in the replay buffer from generating rollouts through the model  makes it possible to increase the number of gradient updates $G$, without overfitting the actors or critics. The following paragraphs will explain the two main components in more detail. Section III-Ba explains the model learning and experience generation. Section III-Bb explains the policy optimization.

\paragraph{Model-Based Learning through MBPO}
\label{par:mbpo}
To learn a world model and generate experience to train the policy on, we adapt the single-agent MBPO algorithm \cite{janner_when_2019} to be suitable for multi-agent domains. Our adaptation uses a centralized world model ($\hat{p}_{\theta}$) that predicts the reward and observations for the next time step given the joint observation of our current time step and the joint action of all agents: $\boldsymbol{o}_{t+1}, r_{t+1} \sim \hat{p}_{\theta}(\boldsymbol{o}_t, \boldsymbol{a}_t)$. 
The remainder of the model architecture, learning, and usage is performed similar to the original implementation. To summarize, we use an ensemble of stochastic neural networks to prevent model bias. The network is trained on the replay buffer as a maximum likelihood estimator for the next observation and rewards.

\paragraph{Policy Optimization through MASAC}
\label{par:masac}
To prevent the instability problems associated with non-stationarity of multi-agent domains, we use a CLDE algorithm for policy optimization. More specifically, we use an Actor-Critic algorithm called MASAC to perform policy optimization based on the experience generated through our world model.

The actor-critic architecture splits up the agent into two separate components: an actor that performs action selection given an observation and a critic that serves as an estimator for the return given a selected joint observation and action. The critic for an agent $i$ is thus a function of the joint observation and joint action of all agents and serves as an estimator for the expected return: $\hat{Q}_{w^i}\left(\boldsymbol{o}_t, \boldsymbol{a}_t\right) \approx Q^{\pi}(s_t,\boldsymbol{a}_t)$. This critic, through taking the joint action of all agents as an input, removes the non-stationarity of the environment.  For this approximation to be a good estimator, the joint observation $\boldsymbol{o}_t$ must contain similar information to the full system state $s_t$. The actor of agent $i$ stochastically selects an action given its individual observation: $a^i_t = \pi_{\phi^i}(\cdot|o_t^i)$. In this article, we consider both the actor and the critic to be neural networks, with parameters $\phi^i$ and $w^i$ respectively.

After every time step, the actors and critics are trained by taking a number of gradient descent steps using a batch of sampled transitions from the replay buffer. In MASAC, the entropy of the policy is maximized simultaneously to maximizing the expected return. This encourages exploration, stabilizing training and improving sample efficiency \cite{gupta_probabilistic_2019, iqbal_actor-attention-critic_2019}. The critic is trained through temporal difference learning, where the target for the critic is set to the following:
\begin{equation}
\begin{split}
&y = r_{t+1} + \gamma \left( \hat{Q}_{w^i}\left(\boldsymbol{o}_{t+1}, \boldsymbol{a}_{t+1}\right) - \alpha \log \pi_{\phi^i}(a_{t+1}^i|{o}_{t+1}^i) \right),\\
&\boldsymbol{a}_{t+1} \sim \boldsymbol{\pi} (\cdot|\boldsymbol{o}_{t+1})
\end{split}
\end{equation}
Without the $\log$ term, this function is equivalent to standard temporal difference learning used in RL. The $\log$ term is an additional term used in (MA) Soft Actor-Critic which rewards policies that have a high entropy. Here, $\alpha$ is a weight that determines the level of entropy regularization. In our implementation, we automatically tune this weight during training by tracking a reference entropy of the policy. The critic's loss function is set to be the mean squared error of this target with the current Q-value estimator:
\begin{equation}
\begin{split}
& \EuScript{L}_{Q^i}\left(w^i, \EuScript{B} \right) = \left(\hat{Q}_{w^i}\left(\boldsymbol{o}_t, \boldsymbol{a}_t\right) - y\right)^2,\\
&(\boldsymbol{o}_t, \boldsymbol{a}_t, r_{t+1}, \boldsymbol{o}_{t+1}) \sim \EuScript{B},  \quad \boldsymbol{a}_{t+1} \sim \boldsymbol{\pi}(\cdot|\boldsymbol{o}_t)
\end{split}
\end{equation}
For the actor, the goal is to maximize the expected return and entropy of the agent's policy. The policy is thus updated through the policy gradient with the following loss function:
\begin{equation}
\begin{split}
&\EuScript{L}_{\pi^i}\left(\phi^i, \EuScript{B} \right) = -\left( \hat{Q}_{w^i}(\boldsymbol{o}_t, \boldsymbol{a}_t) - \alpha \log \pi_{\phi^i}(a_t^i|o_t^i)\right), \\
&(\boldsymbol{o}_t) \sim \EuScript{B}, \quad \boldsymbol{a}_{t} \sim \boldsymbol{\pi}(\cdot|\boldsymbol{o}_t)
\end{split}
\end{equation}

\begin{algorithm}[t]
Initialize actors $\pi_{\phi^i}$, critics $Q_{w^i}$, for each agent $i$\;
Initialize world model $\hat{p}_{\theta}$\; Initialize environment replay buffer $\EuScript{B}_{env}$, model replay buffer $\EuScript{B}_{model}$\;
Initialize environment \;
\For{every timestep $t$}{
Select joint action using $\pi_{\phi^i}$ for each agent $i$\;
Apply joint action in environment using\;
Add transition to  $\EuScript{B}_{env}$ \;
\If{t \% model train frequency = 0}{
Train model $\hat{p}_{\theta}$ on $\EuScript{B}_{env}$ \label{line_start_1}\;
}
\For{M model rollouts}{
Sample $s$ from $\EuScript{B}_{env}$ \;
Perform $k$-step rollout starting from $s$ using $\pi_{\phi^i}$ for each agent $i$ and $\hat{p}_{\theta}$\;
Add transitions to $\EuScript{B}_{model}$ \label{line_end_1}\;
}
\For{G gradient updates}{
\For{every agent $i$}{
Update actor parameters $\phi^i$ using  $\EuScript{B}_{model}$  \label{line_start}\;
Update critic parameters $w^i$ using  $\EuScript{B}_{model}$\label{line_end}\;
}
}
}
\Return{}
\caption{Multi-Agent Model-Based Policy Optimization}
\label{alg:mambpo}
\end{algorithm}

\section{Experiments \& Results}
\label{sec:experiments}
This section tests MAMBPO against MASAC, its model-free counterpart.
The test domains and hyperparameters are detailed in \autoref{subsec:environments} and \autoref{subsec:hyperparameters}, respectively. The results are shown in \autoref{subsec:results}. 


\subsection{Environment Descriptions}
We test using two benchmark domains from the multi-agent particle environments benchmark suite \cite{mordatch_emergence_2018, lowe_multi-agent_2017}, simulating agents in a 2D space.
We use a modified version of the suite \cite{de_witt_deep_2020} that allows continuous actions for each agent: agents output a desired acceleration in the x- and y-direction \footnote{Available at  \url{https://github.com/schroederdewitt/multiagent-particle-envs}. Here, cooperative navigation can be found as \texttt{simple\_spread} and predator-prey can be found as \texttt{simple\_tag\_coop}}.
Each episode consists of 25 time steps.
\label{subsec:environments}
\begin{figure}[ht!]
\centering
\begin{subfigure}{0.23\textwidth}
\centering
  \includegraphics[width=0.8\linewidth]{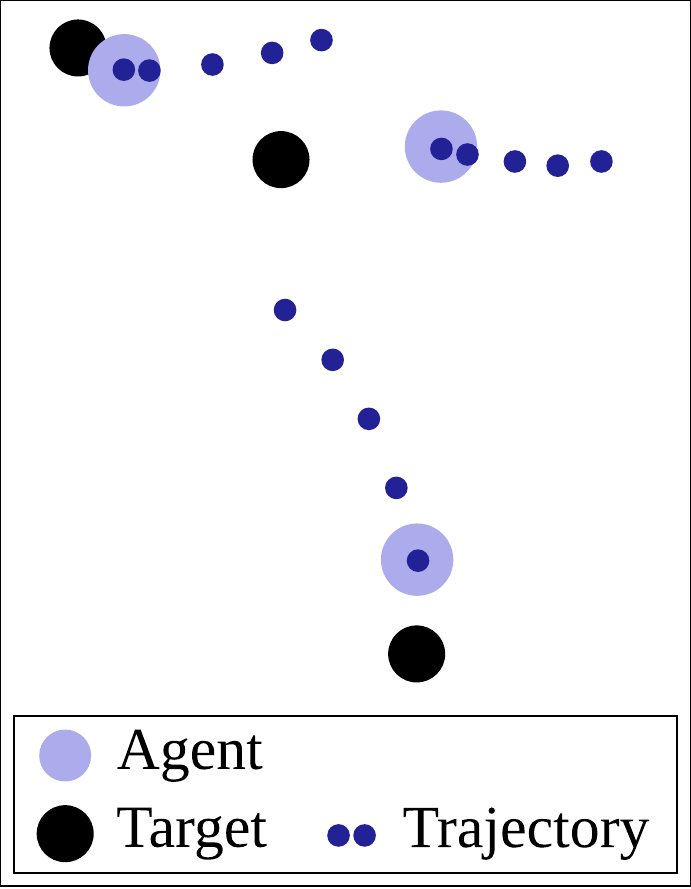}
  \caption{Cooperative navigation}
  \label{fig:env_navigation}
\end{subfigure}
\begin{subfigure}{0.23\textwidth}
\centering
  \includegraphics[width=0.8\linewidth]{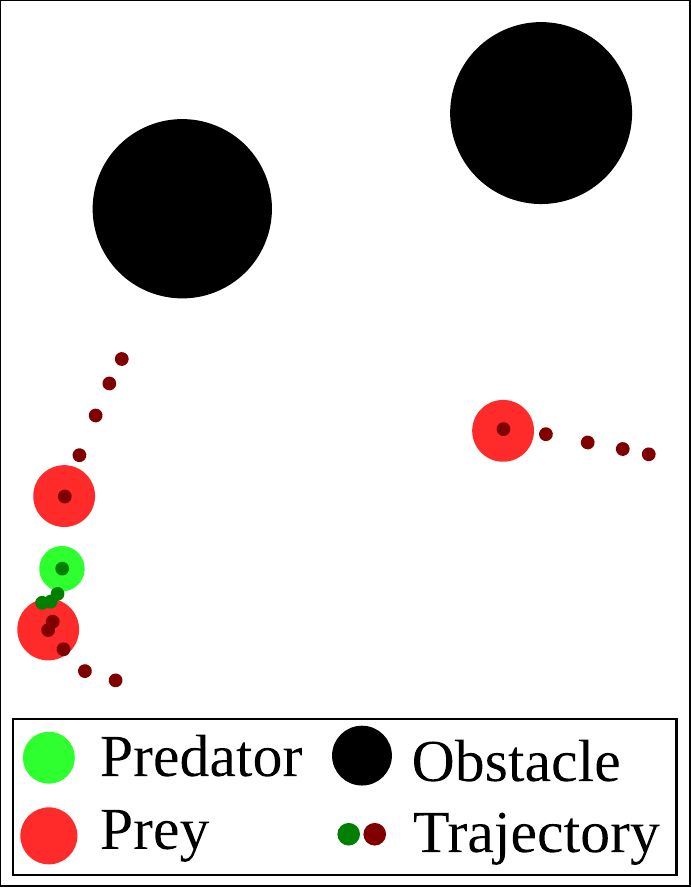}
  \caption{Cooperative predator-prey}
  \label{fig:env_prey}
\end{subfigure}
\caption{Illustrations of the two benchmark domains.}
\end{figure}

\paragraph{Cooperative navigation} In this task, agents need to simultaneously cover three target locations whilst avoiding collisions with each other.
Both the agents and targets are spawned at random locations at the start of each episode.
The agents receive a cooperative reward based on the distance between each target and the closest agent to that target.
The agents perceive the relative locations of all other agents as well as those of all targets.
The domain is illustrated in \autoref{fig:env_navigation}.
This task requires coordination without communication to determine which agent should cover which target.

\paragraph{Cooperative predator-prey} In this task, three collaborating agents need to catch a prey.
This domain is a modification of the simple-tag domain of the multi-agent particle environments benchmark environment suite.
The prey policy is replaced with a fixed heuristic to move away from predators when predators are close.
The predators receive a reward for every time step at least one predator is on top of the prey.
The domain is illustrated in \autoref{fig:env_prey}.
Catching the prey requires less coordination amongst agents compared to the cooperative navigation domain.
However, this domain is of interest because it contains sparse rewards, which can complicate learning.

\subsection{Hyperparameters and Experiment setup}
\label{subsec:hyperparameters}
For a fair comparison between MAMBPO and MASAC, the hyperparameters are kept constant between both environments and both algorithms. 
Both algorithms use the same code, with the exception of a switch that determines whether to train the model $\hat{p}_{\theta}$ and whether to use $\EuScript{B}_{model}$ for training the actors and critics \footnote{All code is available at \url{https://github.com/danielwillemsen/mambpo}.}.
The only hyperparameter that is varied between the two algorithms is the number of gradient steps performed per environment step, as we hypothesize that setting this to a higher value benefits from the larger diversity in the replay buffer in the case of model-based learning.
For a fair comparison, we test the model-free learning algorithm using both the higher number of gradient steps (10) as well as the smaller number of gradient steps (1).
Our low value for gradient steps is the standard value used in single-agent SAC algorithms, but it is still significantly higher than the values used in previous research on these multi-agent domains, as we found this to improve performance for both algorithms.
In MAMBPO, we set the model rollout length to 1 time step.
This was also found to work well in the original MBPO implementation. The full hyperparameter settings can be found in \autoref{tab:fullhyperparameters}.
\begin{table}[t!]
\begin{tabular}{l|l}
Hyperparameter  & Value \\ \hline
Learning rate of [model, actor, critic]            & [0.01, 0.003, 0.003] \\
L2 normalization of model & 0.001\\
Hidden layer sizes [model, actor, critic]  & [4$\times$200, 2$\times$128, 2$\times$256] \\
Batch size [model, actor, critic]  & [512, 256, 256] \\ 
Model training [interval, gradient steps]   & [250, 500] \\
Number of models in ensemble & 10 \\
Actor-Critic target network update rate & 0.01 \\
Initial exploration constant & 0.1 \\
Discount factor & 0.95 \\
Target entropy & $-2$ \\
Min / max log variance of model output & $-5 / -2$ \\
\end{tabular}
\caption{Hyperparameters used in the multi-robot tasks.}
\label{tab:fullhyperparameters}
\end{table}

\begin{figure*}[h!]
\centering
\captionsetup[subfigure]{width=0.95\textwidth}%
\begin{subfigure}[t]{0.48\textwidth}
\centering
  \includegraphics[width=0.9\linewidth]{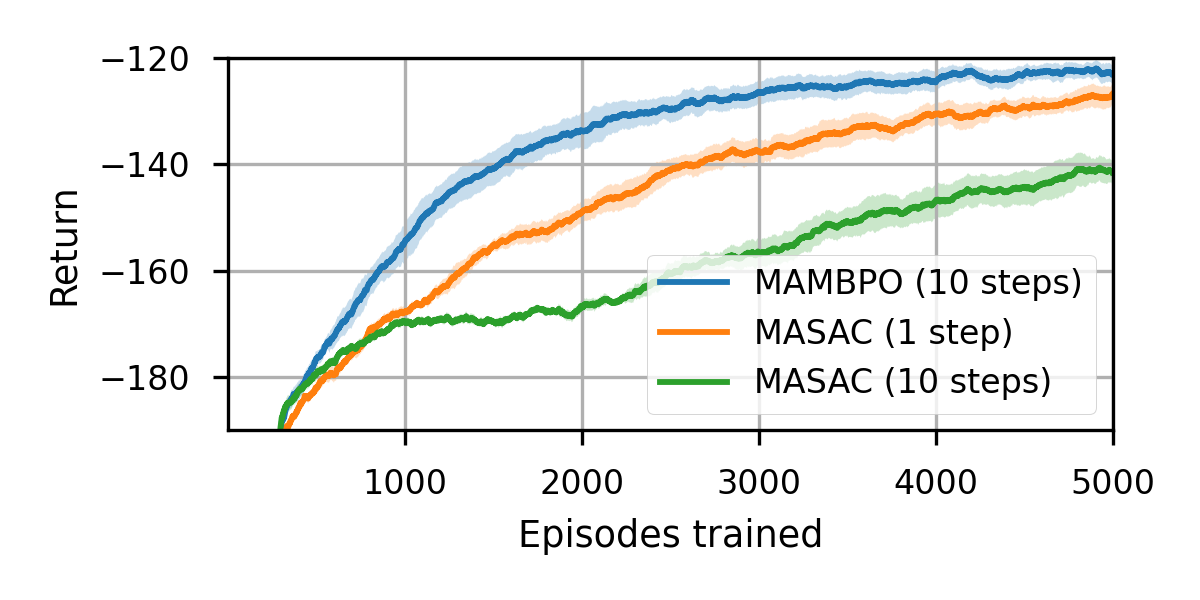}
  \caption{Cooperative navigation. The average cumulative reward per episode of MASAC (1 step) after 5000 training episodes is reached by MAMBPO after only 2980 episodes. This corresponds to about a 1.7 times improvement in sample efficiency.}
\end{subfigure}
\begin{subfigure}[t]{0.48\textwidth}
\centering
  \includegraphics[width=0.9\linewidth]{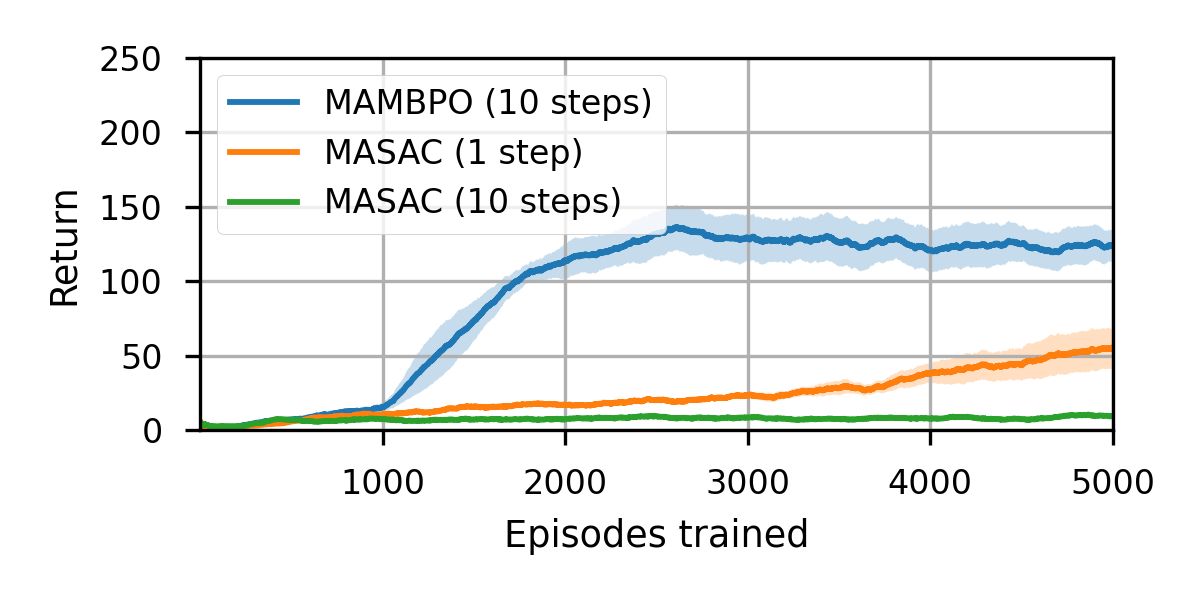}
  \caption{Cooperative predator-prey. The average cumulative reward per episode of MASAC (1 step) after 5000 training episodes is reached by MAMBPO after only 1340 episodes. This corresponds to about a 3.7 times improvement in sample efficiency.}
\end{subfigure}
\caption{Comparison between the performance of MAMBPO and MASAC on the two benchmark tasks. The number of steps indicates the number of gradient steps performed per real environment step. The results are smoothed using a moving average filter with a filter size of 200 episodes. The bold lines and shaded areas indicate the mean and standard error of the mean over 5 independent runs respectively.}
\label{fig:results}
\end{figure*}

\subsection{Results}
\label{subsec:results}

\paragraph{Rewards on benchmark domains}
The results of MAMBPO against MASAC can be seen in \autoref{fig:results} for both domains. MAMBPO results in higher rewards for a given number of episodes, requiring between $1.7 \times$ (cooperative navigation) and $3.7 \times$ (predator-prey) fewer episodes to reach the final score achieved by the standard (1 step) MASAC algorithm.
Increasing the number of gradient steps performed in MASAC does not improve performance or, in the case of predator-prey, prohibits learning altogether. 

\begin{table}[t]
\begin{tabular}{l|cc|cc}
& \multicolumn{2}{c|}{Task 1} & \multicolumn{2}{c}{Task 2}\\
& \multicolumn{2}{c|}{Chance of success} & \multicolumn{2}{c}{Chance of catching prey}\\
Episodes trained & \multicolumn{1}{c|}{MASAC} & MAMBPO & \multicolumn{1}{c|}{MASAC} & MAMBPO \\ \hline
1250                        & \multicolumn{1}{c|}{0.3\%} & 3.2\% &  \multicolumn{1}{c|}{43.5\%} & 72.1\%  \\
2500                        & \multicolumn{1}{c|}{3.8\%} & 20.1\% &  \multicolumn{1}{c|}{55.0\%} & 97.8\% \\
5000                        & \multicolumn{1}{c|}{25.9\%} & 37.1\% &  \multicolumn{1}{c|}{76.3\%} & 98.8\%
\end{tabular}
\caption{Comparison of the chance of covering all three targets for at least 1 timestep, without any collisions occurring during the episode. Evaluated over 5 independent runs and 250 evaluation episodes per run.}
\label{tab:covered}
\end{table}



\begin{figure}[t]
\centering
\captionsetup[subfigure]{width=0.95\textwidth}%
\begin{subfigure}[t]{0.48\linewidth}
\centering
  \includegraphics[width=0.8\linewidth]{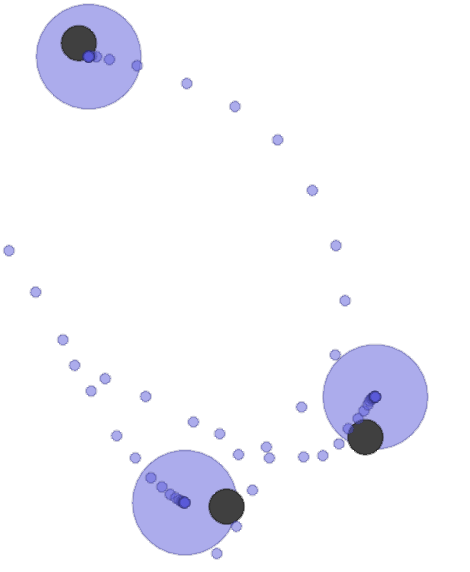}
  \caption{Cooperative navigation, showing a failure case due to a collision (in the 4$^{th}$ time step) between two robots.}
\end{subfigure}
\begin{subfigure}[t]{0.48\linewidth}
\centering
  \includegraphics[width=0.8\linewidth]{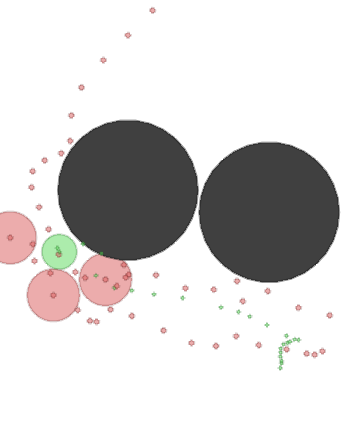}
  \caption{Cooperative predator-prey, showing a successful attempt.}
\end{subfigure}
\caption{Sample trajectories of learned behaviours of the MAMBPO agents after 5000 episodes of training in the two benchmark domains (screenshots from simulator).}
\label{fig:sample_trajectory}
\end{figure}

\paragraph{Success rates on benchmark domains}
For the cooperative navigation task, an episode is successful if all three targets are simultaneously covered for at least 1 time step without any collisions occurring during the episode. We show the success rates at different stages of learning in \autoref{tab:covered}. Although the success rates for both algorithms remain relatively modest (at most $37.1\%$), MAMBPO outperforms MASAC by 16.3 percentage points halfway during training, and 11.2 percentage points after 5000 episodes. Of the unsuccessful episodes still occurring after 5000 episodes of training, $39\%$ is caused by collisions, $36\%$ are caused by not covering all three targets and $25\%$ fail due to both colliding as well as failing to cover all targets. When agents fail to cover all three targets, the agents often do go near all three targets, but not close enough to fully cover them. The high number of collisions might be caused by the relatively low negative reward (-2) associated with them, giving an incentive for agents to go through each other to more quickly reach a targets. For the predator-prey task, we look at the chance of covering the prey by at least one predator for at least one time step during an episode, counting this as ``catching the prey''. The results can be found in \autoref{tab:covered}. In this domain, the success rates are higher. The performance of MAMBPO after only 1250 episodes approaches the performance of MASAC after 5000 episodes.
At 5000 episodes, MAMBPO reaches a success rate of 98.8\%, compared to 76.3\%  with MASAC.
\autoref{fig:sample_trajectory} shows sample trajectories of the learned behaviours.

\begin{figure*}[h!]
\centering
\captionsetup[subfigure]{width=0.95\textwidth}%
\begin{subfigure}[t]{0.48\textwidth}
\centering
  \includegraphics[width=0.9\linewidth]{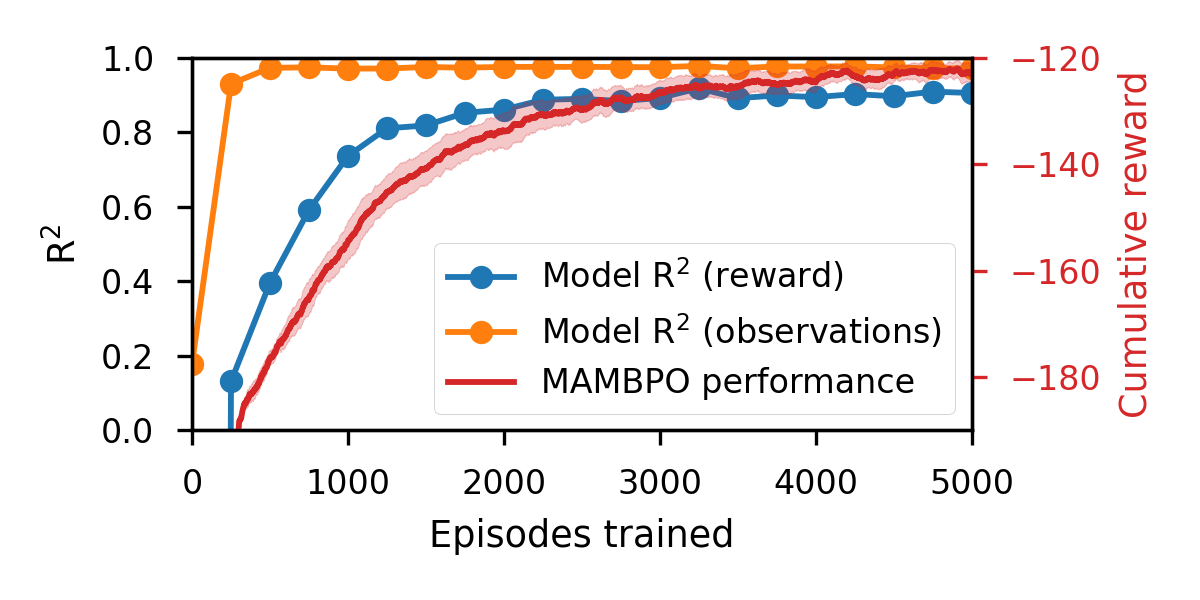}
  \caption{Cooperative navigation. 
  }
\end{subfigure}
\begin{subfigure}[t]{0.48\textwidth}
\centering
  \includegraphics[width=0.9\linewidth]{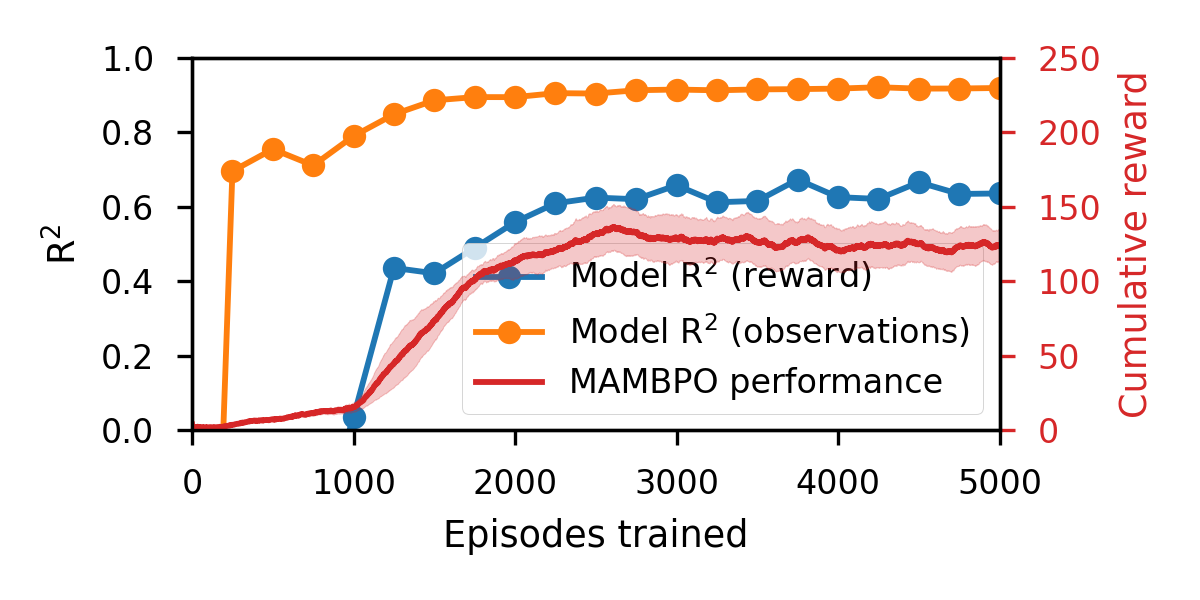}
  \caption{Cooperative predator-prey. 
  }
\end{subfigure}
\caption{Overlay of model quality and agent performance. 
}
\label{fig:model_quality_perf}
\end{figure*}

\paragraph{Model quality}
The performance of agents is overlaid with the quality of the world model in \autoref{fig:model_quality_perf}. We do this by generating a number of trajectories in the environment using the current actor, and measure the mean coefficient of determination ($R^2$) between the single-step predictions of the model and the true single-step rewards and observations.
In both domains, after approximately 2500 episodes of training, the model does not improve further, even though there are still large inaccuracies in the model.  In the cooperative navigation domain, agents continue learning until the end of training, indicating that the model quality is not a bottleneck.
In the predator-prey domain, however, the agents stop improving at a similar point in training as when the model quality stops improving.
This indicates that model-quality might be limiting the final performance of agents. When we investigate the $R^2$ of individual observations rather than the mean $R^2$ of all observations (not shown in the figure), we see that the model performs worse in predicting observations of velocities compared to positions. The velocity of the prey in the predator-prey domain is the most difficult to predict (final $R^2\approx0.60$). This is likely because the actions that the prey takes are not directly visible to the model. Rather, the prey uses a fixed heuristic to move around which is a part of the environment and only the relative velocity of the prey can be observed by the agents themselves. We also find that the model has more difficulty in predicting the reward of the predator-prey domain, compared to the navigation domain. This might be caused by the sparseness and highly non-smooth nature of the rewards in the predator-prey domain. 

Another important aspect of model quality is model bias. As mentioned in the methodology, ensembles of networks were used to reduce the amount of model bias in our algorithm. 
Model bias causes the agents to learn to exploit inaccuracies in the model, leading to lower performance.
The bias can be assessed by comparing the expected rewards of model predictions with real environment rewards. As shown in \autoref{tab:bias}, the differences in rewards are limited, with no substantial evidence of model bias using 1250 samples.
\begin{table}[]
\begin{tabular}{l|lll}
              & \multicolumn{3}{c}{Average Reward ($\pm 2 \times$ standard error of the mean)}                                   \\ \cline{2-4} 
Environment   & \multicolumn{1}{l|}{Real} & \multicolumn{1}{l|}{Predicted} & Bias \\ \hline
Navigation    & \multicolumn{1}{l|}{-4.95 ($\pm$ 0.11)}    & \multicolumn{1}{l|}{-4.94 ($\pm$ 0.11)}      &  - 0.13\%      \\
Predator-prey & \multicolumn{1}{l|}{5.43 ($\pm$ 0.34)}    & \multicolumn{1}{l|}{5.51 ($\pm$ 0.29)}      & + 1.39\%       
\end{tabular}
\caption{Bias of single-step model reward predictions for models trained for 5000 episodes. Each value was calculated based on 5 independent training runs and transitions were sampled from 250 evaluation episodes per training run.}
\label{tab:bias}
\end{table}

\section{Discussion and Conclusion}
\label{sec:conclusion}
The goal of this study is to contribute towards making real-life online learning for groups or even swarms of robots possible, which requires sample-efficient multi-agent RL.
To study the potential benefits of model-based learning on sample efficiency for multi-robot systems, we designed and evaluated a new CLDE algorithm: MAMBPO.
Using this algorithm, we were able to improve the sample efficiency of a model-free baseline, as demonstrated on two benchmark domains, highlighting the potential benefits of model-based learning.

Our tests indicate that model-based learning can improve sample efficiency. The degree of the improvement varies per domain, as is also the case for the single-agent MBPO \cite{janner_when_2019}.
However, although MAMBPO is able to significantly improve sample efficiency, our tests still required thousands of trials to reach desirable performance.
Thus, more improvements are needed to reach our goal of real-life learning for multi-robot systems.
One development would be to improve model learning speed and quality, for example through regularization \cite{kukacka_regularization_2017}) or scalability (for example through attention \cite{iqbal_actor-attention-critic_2019}).
Another development would be to improve the policy optimization approach.
MAMBPO is a model-based version of Multi-Agent Soft-Actor-Critic; yet model-based enhancements can also be explored for other (currently) model-free algorithms such as Attention-Actor-Critic \cite{iqbal_actor-attention-critic_2019} or Q-MIX \cite{rashid_qmix_2018}.
In addition, it should be studied how these algorithms can be scaled up to perform in more complex, real-world tasks.
A solution could be to combine sim-to-real learning \cite{tobin_domain_2017} with model-based RL, such that the agents do not need to start from scratch when acting in the real world.

\bibliography{report}
\end{document}